\documentclass[journal,trans]{IEEEtran}

\usepackage{times}  
\usepackage{helvet}  
\usepackage{courier}  
\usepackage{url}  
\usepackage{graphicx}  

\usepackage{bigstrut}
\usepackage{booktabs}

\usepackage{epsfig}
\usepackage{algorithm} 
\usepackage{algorithmic} 
\usepackage{amsmath}
\usepackage{xcolor}
\usepackage{amsmath,epsfig,graphicx}
\usepackage{amsfonts}
\usepackage{bigstrut,multirow}
\usepackage{threeparttable}

\begin{document}
\title{Unsupervised Semantic-based Aggregation of Deep Convolutional Features}

\author{\IEEEauthorblockN{Jian Xu,
Chunheng Wang,
Chengzuo Qi,
Cunzhao Shi,
and
Baihua Xiao}


}

\markboth{Journal of \LaTeX\ Class Files,~Vol.~14, No.~8, August~2015}%
{Shell \MakeLowercase{\textit{et al.}}: Bare Demo of IEEEtran.cls for IEEE Transactions}

\IEEEtitleabstractindextext{%
\begin{abstract}

In this paper, we propose a simple but effective semantic-based aggregation (SBA) method. The proposed SBA utilizes the discriminative filters of deep convolutional layers as semantic detectors. Moreover, we propose the effective unsupervised strategy to select some semantic detectors to generate the ``probabilistic proposals'', which highlight certain discriminative pattern of objects and suppress the noise of background. The final global SBA representation could then be acquired by aggregating the regional representations weighted by the selected ``probabilistic proposals'' corresponding to various semantic content. Our unsupervised SBA is easy to generalize and achieves excellent performance on various tasks. We conduct comprehensive experiments and show that our unsupervised SBA outperforms the state-of-the-art unsupervised and supervised aggregation methods on image retrieval, place recognition and cloud classification.

\end{abstract}

}

\maketitle
\IEEEdisplaynontitleabstractindextext
\IEEEpeerreviewmaketitle

\section{Introduction}
Over the past decades, feature aggregation has received sustained attention.
Image representations derived by aggregating features such as Scale-Invariant Feature Transform (SIFT)~\cite{sift} and Convolutional Neural Network (CNN)~\cite{feature_map} are shown to be effective for image retrieval~\cite{bow,vlad,fv_cvpr,tri_embed,faemb,rvd,nc,mr,spoc,rmac,crow,interactive,SCDA} and place recognition~\cite{netvlad,CRN}.

Recently, the performance of CNN-based features aggregation methods~\cite{nc,mr,spoc,rmac,crow} rapidly outperforms that of SIFT-based features aggregation methods~\cite{bow,vlad,fv_cvpr,fv_eccv,tri_embed,faemb,rvd}.
Some methods~\cite{off_the_shelf,msop,nc} generate the global representation based on fully connected layer features for image retrieval.
After that, convolutional features are aggregated to obtain the global representation~\cite{mr,spoc,rmac,crow,interactive} and achieve better performance.
Many recent methods~\cite{netvlad,fine_tune_1,fine_tune_2,fine_tune_3,CRN} re-train the image representations end-to-end by collected training datasets.
The fine-tuning process significantly improves the adaptation ability for the  specific  task.
However, these methods~\cite{netvlad,fine_tune_1,fine_tune_2,fine_tune_3,CRN}  need to collect the labeled training datasets and the performance of them heavily relies  on the collected datasets.
The discrepant tasks need different training datasets, for example, the fine-tuned model based on place recognition dataset Pitts250k~\cite{Pittsburgh} is not very suitable for landmark building image retrieval in NetVLAD~\cite{netvlad}.

\begin{figure}
  \centering
  \includegraphics[width=2.5 in]{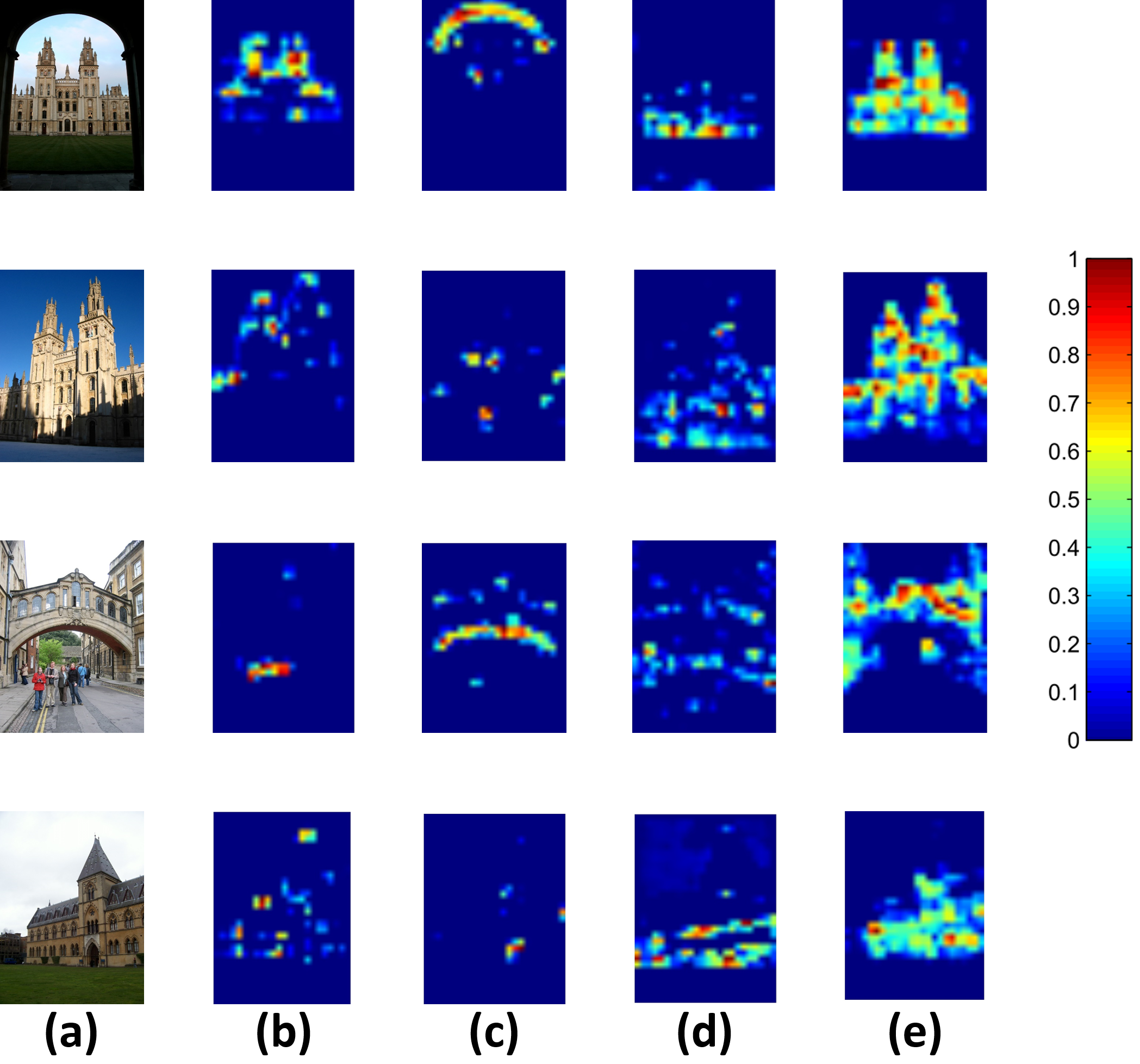}\\
  \caption{Visualization of ``probabilistic proposals''. (a) Some images in Oxford5K~\cite{oxford}. (b)-(e) The various channels of feature maps  in $pool5$ layer from pre-trained VGG16~\cite{VGG}. Each channel of feature maps is activated (warm) by different patterns and some discriminative channels can work as ``probabilistic proposals''.
  }
  \label{feature_map}
\end{figure}

Previous aggregation methods ignore the discriminative information from the object semantic parts.
The semantic-based information is utilized for fine-gained categorization~\cite{filter_response,two_level_attention,unsupervised_part_model,part_based_representation,part_selection_spatial} and the semantic-based representation provides the state-of-the-art performance.
Zhang et al.~\cite{filter_response} pick some distinctive filters which respond to specific patterns significantly and consistently to learn a set of  part detectors. Then, they conditionally  encode the deep filter responses into the final representation based on Fisher vector~\cite{fv_cvpr}.
In recent work~\cite{part_based_representation}, the semantic-based image representation is generated by aggregating selected parts on several different scales.
The recent work~\cite{part_selection_spatial} applies spatial constraints to select part proposals which are generated by selective search~\cite{selective_search}.
Different with these methods, the selected semantic proposals (``probabilistic proposals'') in our algorithm are not constrained to rectangular box but erose shape.

Some recent works~\cite{sppnet,filter_response,visualize} analyze the meaning of feature maps of CNN.
Zeiler et al.~\cite{visualize} show that some input patterns stimulate the special channels of feature maps of the latter convolutional layers.
He et al. visualize the feature maps  generated by some filters of the $conv_{5}$ layer from SPP-net~\cite{sppnet} and show that the filters of deep convolutional layers are activated by specific semantic content and some distinctive filters can work as semantic detectors.
The various channels of convolutional feature maps can represent the  pixel-level label mask of different categories in Fully Convolutional Network (FCN)~\cite{fcn}.
Instance-aware semantic segmentation~\cite{instance2016,instance2017} employs the different channels of shared convolutional layers to detect and segment the various object instance jointly.
Mask R-CNN~\cite{mask_rcnn} demonstrates that the erose proposals perform better than the  rectangular regions on object detection task.
Inspired by above works, we employ some selected discriminative filters of deep convolutional layers as the semantic detectors to generate erose ``probabilistic proposals'', which correspond to  fixed semantic content implicitly.

We define the special channel of normalized feature maps as ``probabilistic proposal'' in this paper.
The ``probabilistic proposal'' encodes the spatial layout of input object's semantic content, and represents the  probability of pixels belonging to fixed semantic.
To further understand  the meanings and characteristics  of the ``probabilistic proposals'', we visualize some images and  corresponding typical ``probabilistic proposals'' in Fig.~\ref{feature_map}.
We select some images in Oxford5K~\cite{oxford} as shown in Fig.~\ref{feature_map} (a).
In Fig.~\ref{feature_map} (b)-(e), we visualize some discriminative channels of feature maps which work as the ``probabilistic proposals''  for the selected images.
Each channel of feature maps is activated (warm) by special patterns corresponding to fixed semantic content and the background is suppressed (cold).
For example, the 220th feature map (Fig.~\ref{feature_map} (b)) of $pool5$ layers from VGG16~\cite{VGG} is most activated by the sharp shape; the 478th feature map  (Fig.~\ref{feature_map} (c)) is most activated by the arc shape; the 483th feature map  (Fig.~\ref{feature_map} (d)) is most activated by the bottom of buildings; the 360th feature map (Fig.~\ref{feature_map} (e)) is most activated by the body of buildings.
We can see that different filters of deep convolutional layers are sensitive to different shapes or semantic, and they highlight different patterns of objects.
Some special patterns of object are discriminative, for example, the 220th feature maps highlight the spire of buildings.
Therefore, filters of deep convolutional layers can work as semantic detectors to pick special patterns corresponding to fixed semantic content.
We select the discriminative filters of deep convolutional layers as the semantic detectors to generate erose ``probabilistic proposals, which are related to different semantic content.

It is gruelling and time-consuming to collect different training datasets for various tasks.
We can try to mine the discriminative information from pre-trained network by employing some unsupervised strategies.
Inspired by the characteristics of feature maps, in this paper we propose a novel and simple way of creating powerful image representation via semantic-based aggregation.
Without need of fine-tuning by different collected training datasets on various tasks, our unsupervised semantic-based aggregation (SBA) method is easy to generalize to different tasks.
We conduct comprehensive experiments on image retrieval~\cite{fine_tune_1,fine_tune_2,fine_tune_3}, place recognition~\cite{netvlad,CRN} and cloud classification~\cite{cloud_cnn}.
Our SBA significantly outperforms most state-of-the-art unsupervised aggregation methods~\cite{mr,spoc,rmac,crow} and supervised methods~\cite{netvlad,fine_tune_1,fine_tune_2,fine_tune_3,CRN}.
Especially on image classification, we improve the non-parameter classification method~\cite{NBNN,C_R_are_one}  based on multi-neighbor information.
In ONE~\cite{C_R_are_one}, Xie et al. demonstrate that the essentials of image classification and retrieval are the same, since both tasks could be tackled by measuring the similarity between images.
We integrate our unsupervised SBA with non-parameter classifier which  requires no learning/training of parameters.

The main contributions of this paper can be summarized as follows:

\subsubsection{``Probabilistic proposal''}
We select some discriminative semantic detectors by succinct unsupervised strategy to generate the ``probabilistic proposals'' corresponding to special semantic content.
Different with previous methods, the selected ``probabilistic proposals'' are not constrained to rectangular box and represent the confidence degree of fixed semantic.
To the best of our knowledge, this paper is the first work to select the erose ``probabilistic proposals'' for image retrieval, and the selected ``probabilistic proposals'' corresponding to special semantic content are tactfully employed to generate  high-dimensional  representation  which contains discriminative semantic information.

\begin{figure*}
  \centering
  \includegraphics[width=6 in]{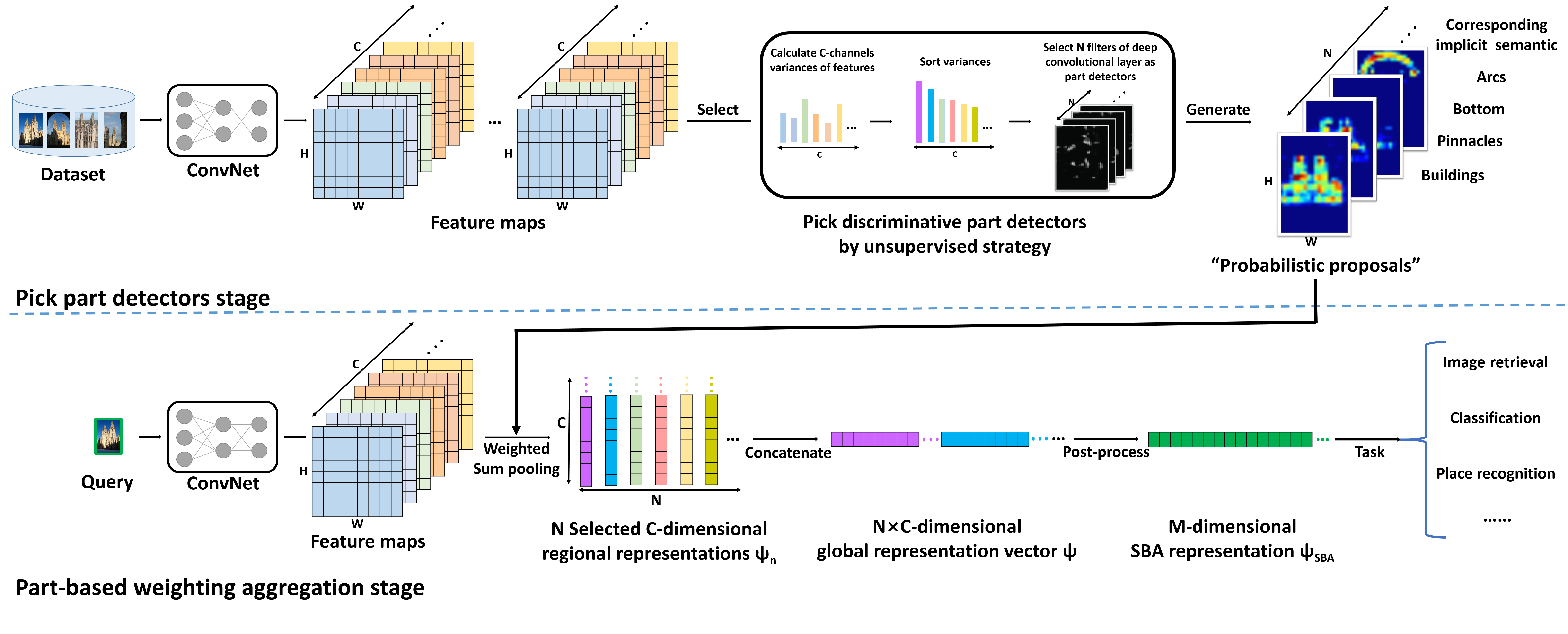}\\
  \caption{ Flow chart of our semantic-based aggregation (SBA) method.
  We pick the discriminative semantic detectors  to generate the ``probabilistic proposals" by the unsupervised strategy in the first off-line stage.
  Each ``probabilistic proposal"  corresponds to fixed semantic content implicitly, such as pinnacles, arcs and bottom of buildings.
  In the aggregation stage, we employ the  selected N ``probabilistic proposals"  to weight and aggregate the feature maps as C-dimensional regional representations, and concatenate N regional representations as the global SBA representation. The final SBA representation can be employed on image retrieval, classification, place recognition and other tasks.
  }
  \label{chart_flow}
\end{figure*}

\subsubsection{Semantic-based aggregation}
We aggregate the convolutional features weighted by selected ``probabilistic proposals'' and concatenate the regional representations as global SBA representation.
Because selected ``probabilistic proposals'' corresponds to fixed semantic but not fixed position, the selected regional representations can be concatenated as the global SBA representation.
Concatenation as the global representation preserves more discrimination than summing regional representations.

\subsubsection{SBA for Classification}
We apply SBA representation on both image classification and retrieval.
We improve ONE~\cite{C_R_are_one} by considering the multi-neighbor relation rather than only nearest-neighbor for image classification.
The non-parameter classifier is easy to generalize to different classification task.
We integrate it with our unsupervised SBA for image classification task, and our holistic  pipeline does not require supervised training.

The updated SBA we present in this paper features several improvements compared to its first version reported in our original conference publication PWA~\cite{PWA}.
Our new version supplements the non-parametric multi-neighbor (MN) classifier for image classification, which is also training-free as our unsupervised SBA. 
We present a more comprehensive experimental evaluation of various tasks (e.g., place recognition and cloud classification) and demonstrate the generalization ability of our SBA.
Finally, we deeply analyze the advantages of our unsupervised SBA in practice.  
We share our code at \url{https://github.com/XJhaoren/PWA}.

\section{Related work}

\subsection{Feature aggregation}
The classical approaches to object based image retrieval involve the use of SIFT features~\cite{sift}. Successful techniques for image retrieval tend to focus on deriving image representations from local descriptors based on aggregation strategy, such as the bag-of-visual-words (BOW) representation~\cite{bow}, BOW with multiple-~\cite{multiple_bow_ijcv,multiple_bow_pami} or soft-assignment~\cite{soft_bow_cvpr,soft_bow_pami}, locality-constrained liner coding~\cite{lclc}, VLAD~\cite{vlad}, Fisher vector~\cite{fv_cvpr,fv_eccv}, triangulation embedding~\cite{tri_embed}, Faemb~\cite{faemb} and robust visual descriptor (RVD) aggregation~\cite{rvd}.

Several recent aggregation methods consider the use of deep CNN fully connected layer features for image retrieval.
Fully connected layer is used as global representation followed by dimensionality reduction~\cite{nc}, and the performance of PCA-compressed representation is better than compact representations computed on traditional SIFT-like features.   Simultaneously, \cite{msop} proposes the more performant representation based on performing orderless VLAD pooling to aggregate the activations of fully connected layers at multiple scale levels. Related to that, the work~\cite{off_the_shelf} reports fairly good retrieval results using sets of multiple sub-patches features of various sizes at different locations that are extracted from fully connected layers of a CNN, without aggregating them into a global representation.

Many recent works derive the visual representation from the activations of convolutional layers.
Razavian et al.~\cite{mr} extend the work~\cite{off_the_shelf} to convolutional layers and the use of convolutional layers leads to much better performance.
After that, the work~\cite{spoc} introduces a compact global image representation based on sum-pooled convolutional features (SPoC) and further shows that the performance of aggregation methods for deep convolutional features is different from shallow features (e.g.,SIFT) because of their higher discriminative ability and different distribution properties.
Recent work~\cite{rmac} proposes a compact image representation derived from the convolutional layer activations which encodes multiple image regions of different sizes without the need to re-feed multiple inputs to network.
Simultaneously, Kalantidis et al.~\cite{crow} extend  the work of~\cite{spoc} by allowing cross-dimensional weighting.

More recently, many works~\cite{netvlad,fine_tune_1,fine_tune_2,fine_tune_3,CRN} that  fine-tune the pre-trained CNN models for image retrieval demonstrate  that the fine-tuned  networks  can bring a significant improvement for image retrieval task.
NetVLAD~\cite{netvlad} plugs a trainable generalized VLAD~\cite{vlad} layer into a CNN and re-trains the model for image retrieval and place recognition via the weakly supervised ranking loss, of which the inputs are the feature maps of convolutional layers and the outputs are the global representations.
After that, the recent works~\cite{fine_tune_1,fine_tune_2,fine_tune_3} fine-tune the deep CNN features for image retrieval.
They aggregate the fine-tuned CNN features and map the global representations based on supervised strategies.
CRN~\cite{CRN} learns the image representation that integrates contextual reweighting of features based on NetVLAD~\cite{netvlad}.
The global representations derived by the supervised strategies  outperform the representations based on pre-trained CNN.
However, these methods~\cite{netvlad,fine_tune_1,fine_tune_2,fine_tune_3,CRN} need to collect the labeled training datasets.
The performance of these methods heavily depends on the collected training datasets.

\subsection{Training-free classification}
Image classification is a fundamental task which is aimed at categorizing images according to their semantic contents.
Recent years, deep convolutional neural networks~\cite{AlexNet,VGG,GoogLeNet,ResNet,AttentionNet} lead to a series breakthroughs for image classification.
However, these learning-based classifiers require an intensive learning/training phase of the classifier parameters.
The non-parametric classifiers base their classification decision directly on the data and require no learning/training of parameters.
Non-parametric classifiers have several very important advantages~\cite{NBNN} that are not shared by most learning-based approaches:
1) Can naturally handle a huge number of classes.
2) Avoid overfitting of parameters, which is a central issue in learning based approaches.
3) Require no learning/training phase. Although training is often viewed as a one-time preprocessing step, re-training of parameters in large dynamic databases may take days or weeks, whereas changing classes/training-sets is instantaneous in non-parametric classifiers.

The most common non-parametric methods rely on nearest-neighbor(NN) distance estimation.
NBNN~\cite{NBNN} employs NN-distances in the space of the local image descriptors without descriptor quantization.
ONE~\cite{C_R_are_one} proposes a unified algorithm for both image classification and retrieval based on nearest-neighbor search of regional representation.
However, these methods only consider nearest-neighbor rather than  multi-neighbor relation.

\section{Aggregation based on ``probabilistic proposals"}
The diagram of the proposed method is shown in Fig.~\ref{chart_flow}.
Based on the dataset, we pick the discriminative semantic detectors  to generate the ``probabilistic proposals" by the unsupervised strategy in the off-line stage.
Each ``probabilistic proposal" corresponds to fixed semantic content implicitly, such as pinnacles, arcs and bottom of buildings.
In the aggregation stage, we employ the  selected N ``probabilistic proposals"  to weight and aggregate the feature maps as C-dimensional regional representations.
We concatenate N regional representations corresponding to special sematic content as the global SBA representation.
After the post-process, SBA representation can be employed on both image retrieval and classification.

In this section, we analyse the characteristics of the filters of deep convolutional layers which can be interpreted as semantic detectors.
We propose the unsupervised strategy to select discriminative semantic detectors to generate  ``probabilistic proposals".
Based on the selected ``probabilistic proposals"  corresponding to special semantic content, we propose a novel and effective SBA aggregation method.
The SBA representation can also be utilized on image classification task based on  non-parametric multi-neighbor (MN) classifier.

We extract  features $f$ from deep convolutional layers  by passing an image $I$ through a pre-trained  or fine-tuned deep network, which consist of $C$ channels feature maps each with height $H$ and width $W$. Finally, the input image $I$ is represented by the aggregated $N\times C$-dimensional vector that are weighted by the  $N$ selected semantic detectors.


\subsection{``Probabilistic proposals"}

\subsubsection{Selection of semantic detectors}
Because the responses  with large variances are significantly different among the various objects, the channels of feature maps with large variances are more discriminative. Therefore, we select semantic detectors according to variances based on dataset.

We first calculate the C-channels variances $V=\{v_{1}, v_{2}, ..., v_{c}, ..., v_{C}\}$  of the $C$-dimensional vectors $g_{i}$ ($i=1,2,...,D$) computed by sum pooling the  $C\times W\times H$-dimensional deep convolutional features $f_{i}$ of image $i$.
\begin{equation}\label{0}
V = \frac{1}{D}\sum\limits_{i = 1}^D {({g_{_i}}}  - \bar g{)^2}
\end{equation}
where $D$ is the  number of database images. $\overline{g}=\frac{1}{D}\sum\limits_{i = 1}^D {g_{i}}$ is the average vector of feature vectors $g_{i}$ ($i=1,2,...,D$).
\begin{equation}\label{0}
{g_{i}} = \sum\limits_{x = 1}^W {\sum\limits_{y = 1}^H {f_{i}(x,y)} }
\end{equation}

Then we sort the variances $\{v_{1}, v_{2}, ..., v_{C}\}$ of C channels.
We select the discriminative deep convolutional layers filters corresponding to large variances as the semantic detectors.
We also observe the filters with large variances to be more discriminative by the following experiment.
We performed retrieval by SBA but we select  (1) 30\% random semantic detectors (2) 30\% semantic detectors with the largest variance. The mAP score for the Oxford5k dataset~\cite{oxford} for (1) is only 0.775$\pm$0.006, which is much small than mAP for (2), 0.790. This verifies that feature maps with large variances are much more discriminative than random feature maps. Moreover, our simple unsupervised selection method not only boosts the performance but also reduces the computational complexity of SBA representation.

\begin{figure*}
  \centering
  \includegraphics[width=7 in]{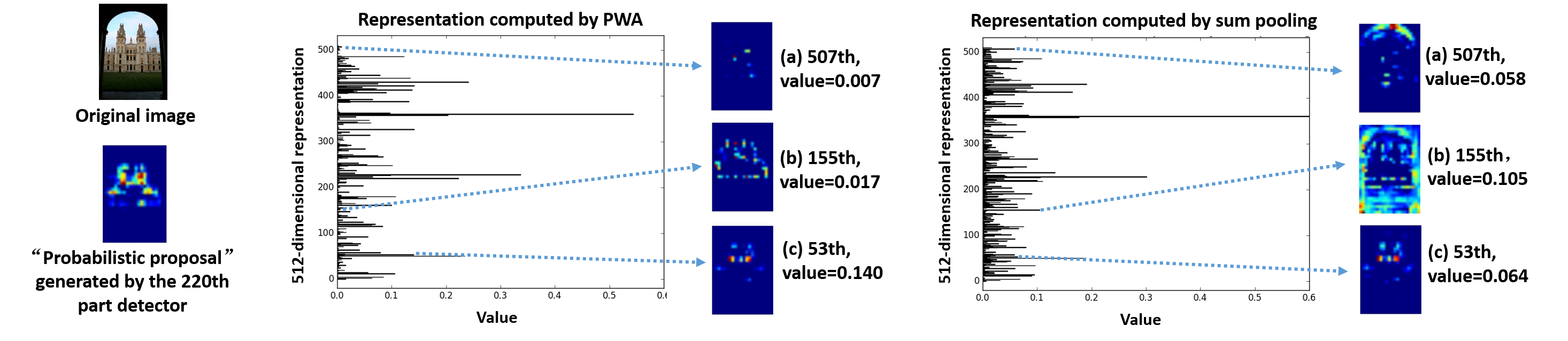}\\
  \caption{The comparison  of the 512-dimensional representations computed by SBA and sum pooling. Weighted by the selected ``probabilistic proposal'',   values of the feature map's channels activated by background (such as (a) 507th and (b) 155th) are reduced. However,  values of the representation  corresponding to similar patterns to the selected ``probabilistic proposal'' (such as (c) 53th) still keep large. The selected ``probabilistic proposal''  suppresses (cold) the  noise of background and highlights (warm)  the special semantic content.}
  \label{sumpooling_pwa}
\end{figure*}

\subsubsection{Effects of ``probabilistic proposals"}

The special channels of feature maps  generated by selected semantic detectors can work as the ``probabilistic proposals" corresponding to fixed semantic content.
To investigate the effects of ``probabilistic proposals" in detail, we compare the 512-dimensional representation computed by sum pooling with the representation weighted by the discriminative ``probabilistic proposals" in Fig.~\ref{sumpooling_pwa}.
As shown in Fig.~\ref{sumpooling_pwa}, the selected ``probabilistic proposal'' generated by 220th semantic detector suppresses the  noise of background and activates the sharp shape.
Weighted by the selected ``probabilistic proposal'', the values of feature maps that are activated by background (such as (a) 507th and (b) 155th) are smaller.
However, the values of the representation  corresponding to similar semantic content  to the selected ``probabilistic proposal'' (such as (c) 53th) still keep large.
As a result, the representations weighted by the discriminative ``probabilistic proposals'' are more discriminative and robust.

Overall, the discriminative filters of latter convolutional layers are interpreted as semantic detectors to generate the ``probabilistic proposals''.
The selected ``probabilistic proposals'' suppress the  noise of background and highlight the discriminative  patterns of objects.
We make use of the selected ``probabilistic proposals'' to weight the activations of convolutional layers and generate the regional representations.
Because each filter of deep convolutional layers activates special pattern, the various selected semantic detectors can be employed to generate the erose proposals  corresponding to special semantic content.
Each proposal corresponds to a fixed semantic pattern  implicitly.
The erose ``probabilistic proposals'' maintain the explicit $W\times H$ object spatial layout which can be addressed naturally by the pixel-to-pixel correspondence provided by convolutions.
Different with R-MAC~\cite{rmac}, the ``probabilistic proposals'' corresponds to fixed semantic content rather than fixed position.
Our ``probabilistic proposals'' are not constrained to box and represent the probability of pixels belong to fixed semantic content.
Although the ``probabilistic proposals''   corresponding to the semantic detectors selected by unsupervised strategy do not explicitly describe the semantic, they implicitly represent discriminative semantic content, such as pinnacles, arcs and bottom of buildings.
Therefore we can concatenate the selected regional representations weighted by  special semantic ``probabilistic proposals'' as the final global representation.
CroW~\cite{crow}, InterActive~\cite{interactive} and SBA can be interpreted as spatial-weighted representations.
InterActive~\cite{interactive} is much more generalized in the aspect of spatial-weighted, which integrates high-level visual context with low-level neuron responses by back-propagation.
Compared to CroW~\cite{crow} and InterActive~\cite{interactive} that sum the spatial-weighted representations, we independently employ the selected semantic detectors to extract the regional representations corresponding to special semantic content and concatenate them as final SBA representation.
The concatenation of regional representations preserves more discriminative information than summation in R-MAC~\cite{rmac}, CroW~\cite{crow} and InterActive~\cite{interactive}.
%

\subsection{SBA design}
In this section, we describe the  SBA  method in detail.
We aggregate the feature maps weighted by the  selected ``probabilistic proposals'' and concatenate the regional representations as global SBA representation.
We reduce the dimensionality of  high-dimensional SBA representation by unsupervised method (PCA) in post-processing.

\subsubsection{Weighted by selected ``probabilistic proposals''} The construction of the SBA representation starts with the weighted sum pooling of the $C\times W\times H$-dimensional deep convolutional features $f$ of image $I$ with height $H$ and width $W$:
\begin{equation}\label{1}
{\psi _{\rm{n}}}(I) = \sum\limits_{x = 1}^W {\sum\limits_{y = 1}^H {{w_n}(x,y)} f(x,y)}
\end{equation}

The coefficients $w_{n}$ are the normalized weights as follows, which depend on the activation values $v_{n}(x,y)$ in position $(x,y)$ of the selected ``probabilistic proposal'' generated by semantic detector $n$:
\begin{equation}\label{2}
{w_n}(x,y) = {\left( {\frac{{{v_n}(x,y)}}{{{{(\sum\limits_{x = 1}^W {\sum\limits_{y = 1}^H {{v_n}} } {{(x,y)}^\alpha })}^{1/\alpha }}}}} \right)^{1/\beta }}
\end{equation}
where $\alpha$ and $\beta$ are parameters of power normalization and power-scaling respectively.

\subsubsection{Concatenation} $N$ selected $C$-dimensional regional representations $\psi_{n}(I)$ are obtained from weighted sum pooling process. We get the global $N\times C$-dimensional representation vector $\psi(I)$ by concatenating selected  regional representations:
\begin{equation}\label{3}
\psi (I) = \left[ {{\psi _1},{\psi _2}, \cdots {\psi _N}} \right]
\end{equation}
where we  select  the N  semantic detectors  depending on the discrimination of them. The selection based on the values of the variances of different  $C$ channels of feature maps both provides  boost in performance  and  enhances the computation efficiency.

\subsubsection{Post-processing} We  perform  $l_{2}$-normalization, PCA compression and whitening on the global representation $\psi(I)$ subsequently and obtain the final M-dimensional representation $\psi_{SBA}(I)$ :
\begin{equation}\label{4}
{\psi _{SBA}}(I) = diag{({\sigma _1},{\sigma _2}, \cdots ,{\sigma _M})^{ - 1}}{V}\frac{{\psi (I)}}{{{{\left. {\left\| {\psi (I)} \right.} \right\|}_2}}}
\end{equation}
where $V$ is the $M\times N$ PCA-matrix, $M$ is the number of the retained dimensionality, and ${\sigma _1},{\sigma _2}, \cdots ,{\sigma _M}$ are the associated singular values.

\subsection{Classification based on multi-neighbor (MN) classifier}
\subsubsection{Distances}
The distances between final M-dimensional SBA representation $\psi_{SBA}(q)$ of query (test) image $(q)$ and SBA representation $\psi_{SBA}(i)$ of database images $(i\in1,2,\cdot\cdot\cdot,D)$ are defined as:
\begin{equation}\label{5}
{{\rm{d}}_i} = \left\| {{\psi _{SBA}}(q) - {\psi _{SBA}}(i)} \right\|_2^2
\end{equation}
where D is the scale of database.

\subsubsection{K nearest-neighbor}
We compute the K near-neighbors $NN_{k}(q)$ of query image $(q)$ based on distances $(d_{1},d_{2},\cdot\cdot\cdot,d_{D})$. Where $k=1,2,\cdot\cdot\cdot,K$ and  $NN_{1}(q)$ is the nearest-neighbors of query image.
The number of neighbors $K$ is the scale of utilized multi-neighbors.

\subsubsection{Confidence scores}
We define the category of the k-th near-neighbors $NN_{k}(q)$ of query image as  $Category(NN_{k}(q))$. The confidence scores $Score_{c}$ for various categories $c$ are computed as follows:

\renewcommand{\algorithmicrequire}{\textbf{Input:}}
\renewcommand{\algorithmicensure}{\textbf{Output:}}
\begin{algorithm} 
\caption{Confidence scores} 
\label{alg1} 
\begin{algorithmic}[1] 
\REQUIRE K near-neighbors $NN_{k}(q)$ of query image.
\ENSURE Confidence scores $Score_{c}$ for various categories $c$.
\STATE $Score_{c}=0, c=1,2,\cdot\cdot\cdot,C$;
\FOR{$k=1$ to $K$}
\STATE $c=Category(NN_{k}(q))$;
\STATE $Score_{c}=Score_{c}+(K-k)$;
\ENDFOR
\end{algorithmic}
\end{algorithm}

\subsubsection{Category prediction}
We select the category corresponding to the maximum of confidence scores $Score_{c}$ as the category prediction finally.

\section{Experiments}
\subsection{Datasets}
We evaluate the performance of SBA and other aggregation algorithms on four standard retrieval datasets (Oxford5k, Paris6k, Oxford105k and Paris106k), Pitts250k~\cite{Pittsburgh} and  SWIMCAT~\cite{SWIMCAT} for image retrieval, place recognition and image classification respectively.

Oxford5k~\cite{oxford} and Paris6k~\cite{paris}  datasets contain photographs collected from Flickr associated with Oxford and Paris landmarks respectively.
The performance is measured using mean average precision (mAP) over the 55 queries  annotated manually. Oxford105k and Paris106k  contain the additional 10,000 distractor images from Flicker~\cite{oxford}.

Pitts250k~\cite{Pittsburgh} contains 250k database images downloaded from Google Street View and 24k test queries generated from Street View but taken at different times, years apart. This dataset is divided into three roughly equal parts for training, validation and testing, each containing around 83k database images and 8k queries, where the division was done geographically to ensure the sets contain independent images in NetVlad~\cite{netvlad}. We compare our SBA with other methods on test set.

SWIMCAT~\cite{SWIMCAT} contains images captured using wide angle high-resolution sky imaging system, a calibrated ground-based WSI designed by ~\cite{WAHRSIS}.
A total of 784 patches comprising five cloud categories are selected from images that were captured in Singapore over the
period January 2013 to May 2014. The five categories include clear sky, patterned clouds, thick dark clouds, thick white clouds, and veil clouds.

\subsection{Implementation details}
We extract deep convolutional features using the pre-trained VGG16~\cite{VGG} and fine-tuned ResNet101 from the work~\cite{fine_tune_3}. In the experiments, Caffe~\cite{caffe} package for CNNs is used. For VGG16 model, we extract convolutional feature maps from the $pool5$ layer and the number of channels is C=512. For ResNet-101 model, we extract convolutional feature maps from the $res5c_{-}relu$ layer and the number of channels is C=2048.
Regarding image size, we keep the original size of the images except for the very large images which are resized to the half size.
The parameters for power normalization and power-scaling are set as $\alpha=2$ and $\beta=2$, throughout our experiments.
The number of multi-neighbor $K$ is set to 40 in MN classifier.

We evaluate the mean average precision (mAP) over the cropped query on image retrieval task.
For fair comparison with the related retrieval methods, we learn the PCA and whitening parameters on Oxford5k when testing on Paris6k and vice versa. 
We follow the standard place recognition evaluation procedure~\cite{netvlad,CRN} on place recognition task.
The query image is deemed correctly localized if at least one of the top $N$ retrieved database images is within 25 meters from the ground truth position of the query.
The percentage of correctly recognized queries (Recall) is then plotted for different value of $N$.
The classification accuracy is reported as the average accuracy of all the test images on cloud classification task.


\begin{table}[htbp]
  \centering
  \caption{Performance of different number of selected semantic detectors (N).
  We aggregate the responses of convolutional layers  by all the C=512 semantic detectors  as the baseline.
  Note, the final representation $\psi_{SBA}(I)$ is reduced  into 4096 dimensionality by PCA.}
    \vspace{0.5em}
    \begin{tabular}{ccc}
    \toprule[1.25pt]
          & \multicolumn{2}{c}{\textbf{Datasets}} \bigstrut\\
\cline{2-3}    \multicolumn{1}{c}{\textbf{N}} & \multicolumn{1}{c}{Oxford5k} & Paris6k \bigstrut\\

    \hline
    \multicolumn{1}{c}{512} & \multicolumn{1}{c}{78.5} & \multicolumn{1}{c}{85.4} \bigstrut[t]\\
    \multicolumn{1}{c}{450} & \multicolumn{1}{c}{78.7} & \multicolumn{1}{c}{85.7} \\
    \multicolumn{1}{c}{350} & \multicolumn{1}{c}{79.0} & \multicolumn{1}{c}{85.9} \\
    \multicolumn{1}{c}{250} & \multicolumn{1}{c}{78.7} & \multicolumn{1}{c}{86.0} \\
    \multicolumn{1}{c}{150} & \multicolumn{1}{c}{79.0} & \multicolumn{1}{c}{85.4} \\
    \multicolumn{1}{c}{50} & \multicolumn{1}{c}{78.2} & \multicolumn{1}{c}{86.1} \\
    \multicolumn{1}{c}{25} & \multicolumn{1}{c}{\textbf{79.1}} & \multicolumn{1}{c}{\textbf{86.1}} \\
    \multicolumn{1}{c}{10} & \multicolumn{1}{c}{77.7} & \multicolumn{1}{c}{83.8} \\
    \bottomrule[1.25pt]
    \end{tabular}%
  \label{select}%
\end{table}%
\subsection{Impact of the parameters}
The main parameters are the numbers  of the selected semantic detectors  and the dimensionality  of final representations $\psi_{SBA}(I)$.

\subsubsection{Select semantic detectors}
We employ the discriminative filters of deep convolutional layers as semantic detectors to generate ``probabilistic proposals''.
The discriminative semantic detectors are selected according to the variances of C channels of feature maps.
We also aggregate the responses of convolutional layers  based on all the C semantic detectors  as the baseline.
We show the results of selecting the first N semantic detectors with the largest variance in Table~\ref{select}.
In this experiment, the final representation $\psi_{SBA}(I)$ is reduced  into 4096 dimensionality by PCA.

The results show that our SBA representation is not heavily relied on the number of selected semantic detectors.
Selecting a small number of semantic detectors (e.g., N=25), we still achieve good performance.
The selection strategy boosts above 0.6\% mAP than  baseline and reduces the computational cost to  1$/$20 of the baseline.
The results demonstrate that our straightforward unsupervised selection strategy is effective.

\begin{table}[htbp]
  \centering
  \caption{Performance of varying dimensionality (M), into which the final representation is reduced.
  The representation is reduced by PCA and whitening.
  Note, we select 25 semantic detectors to aggregate the convolutional features.}
    \vspace{0.5em}
    \begin{tabular}{ccc}
    \toprule[1.25pt]
          & \multicolumn{2}{c}{\textbf{Datasets}} \bigstrut\\
\cline{2-3}    \multicolumn{1}{c}{\textbf{M}} & \multicolumn{1}{c}{Oxford5k} & \multicolumn{1}{c}{Paris6k} \bigstrut\\
    \hline
    \multicolumn{1}{c}{128} & \multicolumn{1}{c}{64.5} & \multicolumn{1}{c}{76.9} \bigstrut[t]\\
    \multicolumn{1}{c}{256} & \multicolumn{1}{c}{68.7} & \multicolumn{1}{c}{79.6} \\
    \multicolumn{1}{c}{512} & \multicolumn{1}{c}{72.0} & \multicolumn{1}{c}{82.3} \\
    \multicolumn{1}{c}{1024} & \multicolumn{1}{c}{75.3} & \multicolumn{1}{c}{84.2} \\
    \multicolumn{1}{c}{2048} & \multicolumn{1}{c}{78.2} & \multicolumn{1}{c}{85.4} \\
    \multicolumn{1}{c}{4096} & \multicolumn{1}{c}{\textbf{79.1}} & \multicolumn{1}{c}{\textbf{86.1}} \bigstrut[b]\\
    \bottomrule[1.25pt]
    \end{tabular}%
  \label{dim}%
\end{table}%

\begin{table*}[htbp]
  \centering
  \caption{Accuracy comparison with the state-of-the-art unsupervised methods. We compare our SBA+QE  with other methods followed by query expansion at the bottom of table. Semantic-based aggregation (SBA) consistently outperforms the state-of-the-art unsupervised aggregation methods.
  }
    \vspace{0.5em}
    \begin{tabular}{rrrrrr}
    \toprule[1.5pt]
          &       & \multicolumn{4}{c}{\textbf{Datasets}} \bigstrut[b]\\
\cline{3-6}    \multicolumn{1}{l}{\textbf{Method}} & \multicolumn{1}{c}{\textbf{Dimensionality}} & \multicolumn{1}{c}{Oxford5k} & \multicolumn{1}{c}{Paris6k} & \multicolumn{1}{c}{Oxford105k} & \multicolumn{1}{c}{Paris106k}  \bigstrut[t]\\
    \toprule[1pt]
    \multicolumn{1}{l}{Tri-embedding~\cite{tri_embed}} & \multicolumn{1}{c}{8k} & \multicolumn{1}{c}{67.6 } & \multicolumn{1}{c}{---} & \multicolumn{1}{c}{61.1 } & \multicolumn{1}{c}{---}  \\
    \multicolumn{1}{l}{FAemb~\cite{faemb}} & \multicolumn{1}{c}{16k} & \multicolumn{1}{c}{70.9 } & \multicolumn{1}{c}{---} & \multicolumn{1}{c}{---} & \multicolumn{1}{c}{---}  \\
    \multicolumn{1}{l}{RVD-W~\cite{rvd}} & \multicolumn{1}{c}{16k} & \multicolumn{1}{c}{68.9 } & \multicolumn{1}{c}{---} & \multicolumn{1}{c}{66.0 } & \multicolumn{1}{c}{---}  \\
    \multicolumn{1}{l}{Razavian et al.~\cite{mr}} & \multicolumn{1}{c}{512} & \multicolumn{1}{c}{46.2 } & \multicolumn{1}{c}{67.4 } & \multicolumn{1}{c}{--- } & \multicolumn{1}{c}{---}  \\
    \multicolumn{1}{l}{Neural Codes~\cite{nc}} & \multicolumn{1}{c}{512} & \multicolumn{1}{c}{43.5 } & \multicolumn{1}{c}{---} & \multicolumn{1}{c}{39.2 } & \multicolumn{1}{c}{---}  \\
    \multicolumn{1}{l}{SPoC~\cite{spoc}} & \multicolumn{1}{c}{256} & \multicolumn{1}{c}{53.1 } & \multicolumn{1}{c}{---} & \multicolumn{1}{c}{50.1 } & \multicolumn{1}{c}{---}  \\
    \multicolumn{1}{l}{InterActive~\cite{interactive}} & \multicolumn{1}{c}{512} & \multicolumn{1}{c}{65.6} & \multicolumn{1}{c}{79.2} & \multicolumn{1}{c}{---} & \multicolumn{1}{c}{---}  \\
    \multicolumn{1}{l}{R-MAC~\cite{rmac}} & \multicolumn{1}{c}{512} & \multicolumn{1}{c}{66.9 } & \multicolumn{1}{c}{83.0 } & \multicolumn{1}{c}{61.6 } & \multicolumn{1}{c}{75.7 }  \\
    \multicolumn{1}{l}{CroW~\cite{crow}} & \multicolumn{1}{c}{512} & \multicolumn{1}{c}{70.8 } & \multicolumn{1}{c}{79.7 } & \multicolumn{1}{c}{65.3 } & \multicolumn{1}{c}{72.2 }  \\
    \hline
    \multicolumn{1}{l}{Previous state-of-the-art} & \multicolumn{1}{c}{} & \multicolumn{1}{c}{70.8 } & \multicolumn{1}{c}{83.0 } & \multicolumn{1}{c}{65.3 } & \multicolumn{1}{c}{75.7 }  \\
    \hline
    \multicolumn{1}{l}{SBA} & \multicolumn{1}{c}{512} & \multicolumn{1}{c}{\textbf{72.0}} & \multicolumn{1}{c}{82.3} & \multicolumn{1}{c}{\textbf{66.2}} & \multicolumn{1}{c}{\textbf{75.8}}  \\
    \multicolumn{1}{l}{SBA} & \multicolumn{1}{c}{1024} & \multicolumn{1}{c}{75.3} & \multicolumn{1}{c}{84.2} & \multicolumn{1}{c}{69.3} & \multicolumn{1}{c}{78.2} \\
    \multicolumn{1}{l}{SBA} & \multicolumn{1}{c}{2048} & \multicolumn{1}{c}{78.2} & \multicolumn{1}{c}{85.4} & \multicolumn{1}{c}{71.1} & \multicolumn{1}{c}{79.7}  \\
    \multicolumn{1}{l}{SBA} & \multicolumn{1}{c}{4096} & \multicolumn{1}{c}{\textbf{79.1}} & \multicolumn{1}{c}{\textbf{86.1}} & \multicolumn{1}{c}{\textbf{73.6}} & \multicolumn{1}{c}{\textbf{80.4}}  \\
    \toprule[1pt]

    \multicolumn{1}{l}{CroW+QE~\cite{crow}} & \multicolumn{1}{c}{512} & \multicolumn{1}{c}{74.9} & \multicolumn{1}{c}{84.8} & \multicolumn{1}{c}{70.6} & \multicolumn{1}{c}{79.4} \bigstrut[b]\\
    \multicolumn{1}{l}{R-MAC+AML+QE~\cite{rmac}} & \multicolumn{1}{c}{512} & \multicolumn{1}{c}{77.3} & \multicolumn{1}{c}{86.5} & \multicolumn{1}{c}{73.2} & \multicolumn{1}{c}{79.8}  \\
    \multicolumn{1}{l}{SBA+QE} & \multicolumn{1}{c}{512} & \multicolumn{1}{c}{{74.8}} & \multicolumn{1}{c}{{86.0}} & \multicolumn{1}{c}{{72.5}} & \multicolumn{1}{c}{{80.7}} \\
    \multicolumn{1}{l}{SBA+QE} & \multicolumn{1}{c}{1024} & \multicolumn{1}{c}{{77.9}} & \multicolumn{1}{c}{{87.8}} & \multicolumn{1}{c}{{76.7}} & \multicolumn{1}{c}{{82.8}} \\
    \multicolumn{1}{l}{SBA+QE} & \multicolumn{1}{c}{2048} & \multicolumn{1}{c}{{80.7}} & \multicolumn{1}{c}{{88.7}} & \multicolumn{1}{c}{{79.3}} & \multicolumn{1}{c}{{83.9}}\\
    \multicolumn{1}{l}{SBA+QE} & \multicolumn{1}{c}{4096} & \multicolumn{1}{c}{\textbf{81.7}} & \multicolumn{1}{c}{\textbf{89.2}} & \multicolumn{1}{c}{\textbf{80.6}} & \multicolumn{1}{c}{\textbf{84.7}} \\
    \toprule[1.5pt]
    \end{tabular}%
  \label{pre-trained}%
\end{table*}%

\subsubsection{Dimensionality reduction}
In order to get shorter representations, we compress the $N\times C$-dimensional  aggregated representation $\psi(I)$ by PCA and whitening process.
Table~\ref{dim} reports the performance of representations with varying dimensionality, M=128 to 4096.
We do not reduce the final representation into higher dimensionality because of  the limited number of images in Oxford5k and Paris6k datasets.
We select N=25 semantic detectors to aggregate the convolutional features in this experiment.

The results show that the performance boosts gradually with the increase of dimensionality and  the best performance is achieved at 4096 dimensionality.
We get the consistent conclusion with other methods,  the compression leads to the loss of discriminative information and performance degradation.
The previous works~\cite{spoc,rmac,crow} aggregate convolutional features as compressed representations with dimensionality under 512, but our SBA representation has more choice of dimensionality.
Compared with~\cite{spoc,rmac,crow}, our SBA methods can generate representations with both low and high dimensionality and achieve better performance on most datasets.
The dimensionality of SBA representation can be chosen according to the tradeoff between performance and efficiency  on different tasks.

\subsection{Comparison with the state-of-the-art}
\begin{table*}[htbp]
  \centering
  \caption{Accuracy comparison with the state-of-the-art supervised methods. Employing the convolutional layer features of fine-tuned network~\cite{fine_tune_3}, we achieve the comparable performance with the state-of-the-art methods with end-to-end supervised training.}
    \vspace{0.5em}
    \begin{tabular}{rrrrrr}
     \toprule[1.25pt]
          &       & \multicolumn{4}{c}{\textbf{Datasets}} \bigstrut\\
\cline{3-6}    \multicolumn{1}{l}{\textbf{Method}} & \multicolumn{1}{c}{\textbf{Dimensionality}} & \multicolumn{1}{c}{Oxford5k} & \multicolumn{1}{c}{Paris6k} & \multicolumn{1}{c}{Oxford105k} & \multicolumn{1}{c}{Paris106k} \bigstrut[t]\\
    \toprule[1pt]
\multicolumn{1}{l}{NetVLAD~\cite{netvlad}} & \multicolumn{1}{c}{512} & \multicolumn{1}{c}{67.6} & \multicolumn{1}{c}{74.9} &\multicolumn{1}{c}{---} & \multicolumn{1}{c}{---}  \\
    \multicolumn{1}{l}{NetVLAD~\cite{netvlad}} & \multicolumn{1}{c}{2048} & \multicolumn{1}{c}{70.8} & \multicolumn{1}{c}{78.3} &\multicolumn{1}{c}{---} & \multicolumn{1}{c}{---}  \\
  \multicolumn{1}{l}{CRN~\cite{CRN}} & \multicolumn{1}{c}{512} & \multicolumn{1}{c}{64.5} & \multicolumn{1}{c}{---} &\multicolumn{1}{c}{62.2} & \multicolumn{1}{c}{---}  \\
    \multicolumn{1}{l}{CRN~\cite{CRN}} & \multicolumn{1}{c}{2048} & \multicolumn{1}{c}{68.3} & \multicolumn{1}{c}{---} &\multicolumn{1}{c}{66.2} & \multicolumn{1}{c}{---}  \\
    \multicolumn{1}{l}{CNNBoW~\cite{fine_tune_1}} & \multicolumn{1}{c}{512} & \multicolumn{1}{c}{79.7} & \multicolumn{1}{c}{83.8} &\multicolumn{1}{c}{73.9} & \multicolumn{1}{c}{76.4}  \\

    \multicolumn{1}{l}{DeepRepresentation~\cite{fine_tune_3}} & \multicolumn{1}{c}{2048} & \multicolumn{1}{c}{86.1} & \multicolumn{1}{c}{94.5} & \multicolumn{1}{c}{82.8} & \multicolumn{1}{c}{90.6}  \bigstrut[b]\\
    \hline
    \multicolumn{1}{l}{Previous state-of-the-art} & \multicolumn{1}{l}{} & \multicolumn{1}{c}{86.1} & \multicolumn{1}{c}{94.5} &\multicolumn{1}{c}{82.8} & \multicolumn{1}{c}{90.6}  \bigstrut\\
 \hline
    \multicolumn{1}{l}{SBA (with pre-trained VGG16)} & \multicolumn{1}{c}{512} & \multicolumn{1}{c}{\textbf{72.0}} & \multicolumn{1}{c}{\textbf{82.3}} &\multicolumn{1}{c}{\textbf{66.2}} & \multicolumn{1}{c}{\textbf{75.8}} \bigstrut[t]\\
 \multicolumn{1}{l}{SBA (with pre-trained VGG16)} & \multicolumn{1}{c}{2048} & \multicolumn{1}{c}{\textbf{78.2}} & \multicolumn{1}{c}{\textbf{85.4}} &\multicolumn{1}{c}{\textbf{71.1}} & \multicolumn{1}{c}{\textbf{79.7}} \bigstrut[t]\\
    \multicolumn{1}{l}{SBA (with fine-tuned ResNet101)} & \multicolumn{1}{c}{2048} & \multicolumn{1}{c}{\textbf{87.8}} & \multicolumn{1}{c}{\textbf{94.9}} &\multicolumn{1}{c}{\textbf{82.8}} & \multicolumn{1}{c}{\textbf{91.0}} \bigstrut[t]\\
    \toprule[1.25pt]
    \end{tabular}%
  \label{fine-tuned}%
\end{table*}%

\subsubsection{Image retrieval}

In the first part of Table~\ref{pre-trained}, we compare our SBA  method using pre-trained VGG16~\cite{VGG} with the state-of-the-art unsupervised methods, which employ global representations of images.  Our SBA representation significantly outperform them on all four standard retrieval datasets.
In particular, the gain is more than 8.3\% in mAP on Oxford5k and Oxford105k datasets.
The results  demonstrate that our SBA representation weighted by the selected ``probabilistic proposals'' is effective  and  discriminative for image retrieval.
Our 512-dimensional SBA representation is comparable with the previous state-of-the-art, and its results are only lower than R-MAC~\cite{rmac} on Paris6k.
The SBA representation with higher dimensionality (such as 1024, 2048 and 4096)  consistently outperform all of them on all datasets.

We compare other methods that contain query  expansion (QE) and spatial verification stages with our approach in the second part of Table~\ref{pre-trained}.
In the experiments, we use average query expansion (QE)~\cite{qe}  computed by the top 10 query results.
Our SBA+QE method performs better than the related works~\cite{rmac,crow} on all datasets.
Although the approximate max pooling localization  (AML) process in R-MAC~\cite{rmac} requires a costly verification stage and the extra memory storage, our SBA+QE still achieves better performance than R-MAC+AML+QE.
We also compare our  method  with the current state-of-the-art supervised methods  containing end-to-end training process \cite{netvlad,CRN,fine_tune_1,fine_tune_3}) in Table~\ref{fine-tuned}.
In order to compare with them, we employ convolutional layers features of fine-tuned ResNet101 from the work~\cite{fine_tune_3} in SBA (with fine-tuned ResNet101).
Because these methods~\cite{fine_tune_1,fine_tune_3} map the final representation by the supervised methods for similarity evaluation, we also map the SBA representations for comparison purposes.
In order to keep consistently unsupervised, we utilize the unsupervised IME layer~\cite{IME} to map our SBA representations for similarity evaluation.

The results show that our unsupervised SBA representation outperforms the state-of-the-art supervised methods~\cite{netvlad,CRN,fine_tune_1,fine_tune_3} on all datasets.
Furthermore, the effectiveness of the supervised methods  is heavily relied on the collected training set.
The training dataset (Pitts250k~\cite{Pittsburgh} or San Francisco~\cite{SanFrancisco}) in NetVLAD~\cite{netvlad} and CRN~\cite{CRN} is not very suitable for Oxford5k and Paris6k datasets, so their performance on Oxford5k and Paris6k datasets improves unsignificantly after re-training.
However, our unsupervised SBA method can make better use of the convolutional features extracted from both pre-trained and fine-tuned CNN model to represent the images and does not need the further supervised re-training.
Even employing convolutional layer feature of pre-trained VGG~\cite{VGG}, our unsupervised SBA (with pre-trained VGG16)  still achieves better performance than supervised NetVLAD~\cite{netvlad} and CRN~\cite{CRN} on all datasets as shown in Table~\ref{fine-tuned}.
Considering the fact that the annotated training dataset is difficult to collect, it is impractical to fine-tune the model for each discrepant task respectively.
Our unsupervised SBA method is very suitable for this condition.
Our SBA method retains more discriminative  information  of the retrieval object and significantly suppress the noise of background, and  better utilizes the convolutional features extracted from both pre-trained and fine-tuned CNN models.

\subsubsection{Place recognition}
To assess benefits of our approach we compare our unsupervised SBA representations against both unsupervised and supervised state-of-the-art methods on place recognition.
The baseline either use max-pooling or aggregate the descriptors into VLAD, but perform no further task-special training.
The state-of-the-art local feature based compact descriptor is consists of VLAD pooling~\cite{vlad} with intra-normalization~\cite{all_about_VLAD} on top of densely extracted RootSIFTs~\cite{sift,three_thing}. The descriptor is optionally reduced to 4096 dimensions using PCA (learnt on the training set) combined with whitening and $l_{2}$-normalization~\cite{whitening}.
The state-of-the-art supervised NetVLAD~\cite{netvlad} and CRN~\cite{CRN} with whitening based on VGG16 are re-trained by both Pitts250k~\cite{Pittsburgh} training set and TokyoTM~\cite{netvlad}.

\begin{figure}
  \centering
  \includegraphics[width=3 in]{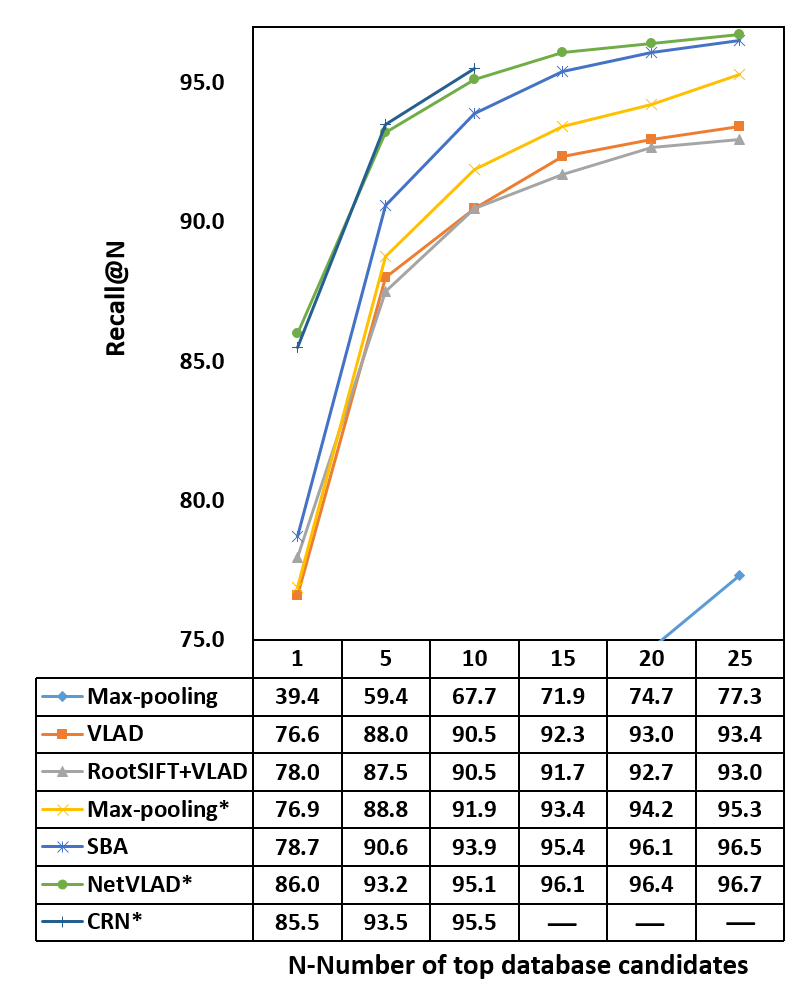}\\
  \caption{Comparison of our SBA versus baselines and state-of-the-art on place recognition.
 Methods marked with an * are re-trained by collected training dataset.
In order to displays the comparison clearly, we only show the recall higher than 75$\%$ in the line chart.
  }
  \label{place_recognition}
\end{figure}

Fig.~\ref{place_recognition} shows that our unsupervised SBA significantly outperforms RootSIFT+VLAD and therefore sets the state-of-the-art for unsupervised compact descriptors on all benchmarks.
Furthermore we also achieve better performance than the re-trained sum-pooling.
Our percentage of correctly recognized queries (Recall) is only lower than supervised NetVLAD~\cite{netvlad} and CRN~\cite{CRN} that re-train the pre-trained network and NetVLAD aggregation layer end-to-end by collected training datasets.

\subsubsection{Cloud classification}
In order to demonstrate the effectiveness and generalization of our unsupervised SBA representation, we apply it on cloud classification which is weakly related to the network pre-trained by Imagenet~\cite{ImageNet}.
We compare the proposed SBA representation with the following methods:
1) the 12-D Heinle feature~\cite{Heinle} which captures color, edge, and texture information of a sky/cloud image;
2) LBP~\cite{LBP}, which is very effective for describing textures and has been widely used for cloud representation;
3) SDCAF~\cite{DCAF} which employs sum-pooling to aggregate the deep convolutional features of CNN for cloud classification.
The parameters of LBP $(P, R)=(8, 1), (16, 2), (24, 3)$ are tested, where $P$ is the number of sampling points on a circle of radius $R$. We only list the results of $(P, R)=(16, 2)$ with which the best performance is achieved.

The classification accuracy of Heinle feature~\cite{Heinle,DCAF}, LBP~\cite{LBP,DCAF}, SDCAF~\cite{DCAF} and proposed SBA+MN are 85.26, 90.26, 96.40 and 97.83 respectively, where Heinle feature~\cite{Heinle,DCAF}, and LBP~\cite{LBP,DCAF}  are tested with linear SVM~\cite{LIBLINEAR}.
The results demonstrate that our unsupervised SBA and no-parametric MN classifier requiring no training of parameters are effective for image classification and achieve near-perfect performance.
Without re-training, they are easy to transfer to various classification objects.
The accuracy of SBA+NN is 96.68, which is about 1.2  percent lower than SBA+MN.
The comparison shows that multi-neighbor information is favorable and effective.

\subsubsection{Results analysis}
Our unsupervised SBA and non-parametric multi-neighbor (MN) classifier are both training-free of the annotated dataset.
The comprehensive experiments of various tasks demonstrate the generalization ability of our SBA.
Our method is well suited for the situation where the annotated training dataset is gruelling and time-consuming to collect.
Due to the fact that the performance of re-trained methods is heavily relied on the quality of collected training datasets, it is another choice to make better use of the discriminative features extracted form pre-trained network by unsupervised strategy like our method in practice.

\section{Conclusion}
In this paper, we propose a novel semantic-based aggregation (SBA) method.
The key  characteristic of our method is that it employs discriminative semantic detectors selected by unsupervised strategy to generate ``probabilistic proposals''.
Based on the selected ``probabilistic proposals''  corresponding to special semantic content implicitly, we weight and aggregate the  deep convolutional features  extracted from pre-trained or fine-tuned CNN models.
Due to the selected ``probabilistic proposals''  corresponding to fixed semantic content but not fixed position, we concatenate the regional representations as global SBA representation.
The results show that our SBA representation suppress the noise of background and highlight the discriminative patterns of retrieval objects.

Experiments on standard image retrieval, place recognition and cloud classification datasets demonstrate that our unsupervised approach outperforms most previous state-of-the-art unsupervised and supervised aggregation methods.
It is worth noting that our unsupervised SBA method is very suitable and effective for the situation where the annotated training dataset is difficult to collect.



\section{Acknowledgments}
This work was supported by the National Natural Science Foundation of China under Grant 61531019, Grant 61601462, and Grant 71621002.

%

\ifCLASSOPTIONcaptionsoff
  \newpage
\fi

\bibliographystyle{IEEEtran}
\bibliography{mybib}

\begin{thebibliography}{10}
\providecommand{\url}[1]{#1}
\csname url@samestyle\endcsname
\providecommand{\newblock}{\relax}
\providecommand{\bibinfo}[2]{#2}
\providecommand{\BIBentrySTDinterwordspacing}{\spaceskip=0pt\relax}
\providecommand{\BIBentryALTinterwordstretchfactor}{4}
\providecommand{\BIBentryALTinterwordspacing}{\spaceskip=\fontdimen2\font plus
\BIBentryALTinterwordstretchfactor\fontdimen3\font minus
  \fontdimen4\font\relax}
\providecommand{\BIBforeignlanguage}[2]{{%
\expandafter\ifx\csname l@#1\endcsname\relax
\typeout{** WARNING: IEEEtran.bst: No hyphenation pattern has been}%
\typeout{** loaded for the language `#1'. Using the pattern for}%
\typeout{** the default language instead.}%
\else
\language=\csname l@#1\endcsname
\fi
#2}}
\providecommand{\BIBdecl}{\relax}
\BIBdecl

\bibitem{sift}
D.~G. Lowe, ``Distinctive image features from scale-invariant keypoints,''
  \emph{International Journal of Computer Vision}, vol.~60, no.~60, pp.
  91--110, 2004.

\bibitem{feature_map}
Y.~LeCun, B.~Boser, J.~S. Denker, D.~Henderson, R.~E. Howard, W.~Hubbard, and
  L.~D. Jackel, ``Backpropagation applied to handwritten zip code
  recognition,'' \emph{Neural computation}, vol.~1, no.~4, pp. 541--551, 1989.

\bibitem{bow}
J.~Sivic and A.~Zisserman, ``Video google: A text retrieval approach to object
  matching in videos,'' in \emph{IEEE International Conference on Computer
  Vision}, 2003, p. 1470.

\bibitem{vlad}
H.~Jegou, F.~Perronnin, M.~Douze, J.~Sanchez, P.~Perez, and C.~Schmid,
  ``Aggregating local image descriptors into compact codes,'' \emph{IEEE
  transactions on pattern analysis and machine intelligence}, vol.~34, no.~9,
  pp. 1704--1716, 2012.

\bibitem{fv_cvpr}
F.~Perronnin and C.~Dance, ``Fisher kernels on visual vocabularies for image
  categorization,'' in \emph{IEEE Conference on Computer Vision and Pattern
  Recognition}, 2007, pp. 1--8.

\bibitem{tri_embed}
H.~Gou and A.~Zisserman, ``Triangulation embedding and democratic aggregation
  for image search,'' in \emph{Computer Vision and Pattern Recognition}, 2014,
  pp. 3310--3317.

\bibitem{faemb}
T.-T. Do, Q.~D. Tran, and N.-M. Cheung, ``Faemb: a function approximation-based
  embedding method for image retrieval,'' in \emph{IEEE Conference on Computer
  Vision and Pattern Recognition}, 2015, pp. 3556--3564.

\bibitem{rvd}
S.~S. Husain and M.~Bober, ``Improving large-scale image retrieval through
  robust aggregation of local descriptors,'' \emph{IEEE Transactions on Pattern
  Analysis and Machine Intelligence}, 2016.

\bibitem{nc}
A.~Babenko, A.~Slesarev, A.~Chigorin, and V.~Lempitsky, ``Neural codes for
  image retrieval,'' in \emph{European conference on computer vision}.\hskip
  1em plus 0.5em minus 0.4em\relax Springer, 2014, pp. 584--599.

\bibitem{mr}
A.~S. Razavian, J.~Sullivan, S.~Carlsson, and A.~Maki, ``Visual instance
  retrieval with deep convolutional networks,'' \emph{ITE Transactions on Media
  Technology and Applications}, vol.~4, no.~3, pp. 251--258, 2016.

\bibitem{spoc}
A.~Babenko and V.~Lempitsky, ``Aggregating local deep features for image
  retrieval,'' in \emph{IEEE international conference on computer vision},
  2015, pp. 1269--1277.

\bibitem{rmac}
G.~Tolias, R.~Sicre, and H.~J¨¦gou, ``Particular object retrieval with integral
  max-pooling of cnn activations,'' \emph{ICLR}, 2016.

\bibitem{crow}
Y.~Kalantidis, C.~Mellina, and S.~Osindero, ``Cross-dimensional weighting for
  aggregated deep convolutional features,'' in \emph{European Conference on
  Computer Vision}.\hskip 1em plus 0.5em minus 0.4em\relax Springer, 2016, pp.
  685--701.

\bibitem{interactive}
L.~Xie, L.~Zheng, J.~Wang, A.~L. Yuille, and Q.~Tian, ``Interactive:
  Inter-layer activeness propagation,'' in \emph{IEEE Conference on Computer
  Vision and Pattern Recognition}, 2016, pp. 270--279.

\bibitem{SCDA}
X.-S. Wei, J.-H. Luo, J.~Wu, and Z.-H. Zhou, ``Selective convolutional
  descriptor aggregation for fine-grained image retrieval,'' \emph{IEEE
  Transactions on Image Processing}, vol.~26, no.~6, pp. 2868--2881, 2017.

\bibitem{netvlad}
R.~Arandjelovic, P.~Gronat, A.~Torii, T.~Pajdla, and J.~Sivic, ``Netvlad: Cnn
  architecture for weakly supervised place recognition,'' in \emph{IEEE
  Conference on Computer Vision and Pattern Recognition}, 2016, pp. 5297--5307.

\bibitem{CRN}
H.~J. Kim, E.~Dunn, and J.-M. Frahm, ``Learned contextual feature reweighting
  for image geo-localization,'' in \emph{IEEE Conference on Computer Vision and
  Pattern Recognition}, 2017.

\bibitem{fv_eccv}
F.~Perronnin, J.~Nchez, and T.~Mensink, ``Improving the fisher kernel for
  large-scale image classification,'' in \emph{European Conference on Computer
  Vision}, 2010, pp. 143--156.

\bibitem{off_the_shelf}
A.~Sharif~Razavian, H.~Azizpour, J.~Sullivan, and S.~Carlsson, ``Cnn features
  off-the-shelf: an astounding baseline for recognition,'' in \emph{IEEE
  Conference on Computer Vision and Pattern Recognition Workshops}, 2014, pp.
  806--813.

\bibitem{msop}
Y.~Gong, L.~Wang, R.~Guo, and S.~Lazebnik, ``Multi-scale orderless pooling of
  deep convolutional activation features,'' in \emph{European conference on
  computer vision}.\hskip 1em plus 0.5em minus 0.4em\relax Springer, 2014, pp.
  392--407.

\bibitem{fine_tune_1}
F.~Radenovic, G.~Tolias, and O.~Chum, ``Cnn image retrieval learns from bow:
  Unsupervised fine-tuning with hard examples,'' in \emph{European Conference
  on Computer Vision}.\hskip 1em plus 0.5em minus 0.4em\relax Springer, 2016,
  pp. 3--20.

\bibitem{fine_tune_2}
A.~Gordo, J.~Almazan, J.~Revaud, and D.~Larlus, ``Deep image retrieval:
  Learning global representations for image search,'' in \emph{European
  Conference on Computer Vision}.\hskip 1em plus 0.5em minus 0.4em\relax
  Springer, 2016, pp. 241--257.

\bibitem{fine_tune_3}
------, ``End-to-end learning of deep visual representations for image
  retrieval,'' \emph{International Journal of Computer Vision}, pp. 1--18,
  2016.

\bibitem{Pittsburgh}
A.~Torii, J.~Sivic, T.~Pajdla, and M.~Okutomi, ``Visual place recognition with
  repetitive structures,'' in \emph{IEEE conference on computer vision and
  pattern recognition}, 2013, pp. 883--890.

\bibitem{oxford}
J.~Philbin, O.~Chum, M.~Isard, J.~Sivic, and A.~Zisserman, ``Object retrieval
  with large vocabularies and fast spatial matching,'' in \emph{IEEE Conference
  on Computer Vision and Pattern Recognition}.\hskip 1em plus 0.5em minus
  0.4em\relax IEEE, 2007, pp. 1--8.

\bibitem{VGG}
K.~Simonyan and A.~Zisserman, ``Very deep convolutional networks for
  large-scale image recognition,'' \emph{ICLR}, 2015.

\bibitem{filter_response}
X.~Zhang, H.~Xiong, W.~Zhou, W.~Lin, and Q.~Tian, ``Picking deep filter
  responses for fine-grained image recognition,'' in \emph{IEEE Conference on
  Computer Vision and Pattern Recognition}, 2016, pp. 1134--1142.

\bibitem{two_level_attention}
T.~Xiao, Y.~Xu, K.~Yang, J.~Zhang, Y.~Peng, and Z.~Zhang, ``The application of
  two-level attention models in deep convolutional neural network for
  fine-grained image classification,'' in \emph{IEEE Conference on Computer
  Vision and Pattern Recognition}, 2015, pp. 842--850.

\bibitem{unsupervised_part_model}
M.~Simon and E.~Rodner, ``Neural activation constellations: Unsupervised part
  model discovery with convolutional networks,'' in \emph{IEEE International
  Conference on Computer Vision}, 2015, pp. 1143--1151.

\bibitem{part_based_representation}
Y.~Zhang, X.-S. Wei, J.~Wu, J.~Cai, J.~Lu, V.-A. Nguyen, and M.~N. Do, ``Weakly
  supervised fine-grained categorization with part-based image
  representation,'' \emph{IEEE Transactions on Image Processing}, vol.~25,
  no.~4, pp. 1713--1725, 2016.

\bibitem{part_selection_spatial}
X.~He and Y.~Peng, ``Weakly supervised learning of part selection model with
  spatial constraints for fine-grained image classification.'' in \emph{AAAI},
  2017, pp. 4075--4081.

\bibitem{selective_search}
J.~R. Uijlings, K.~E. Van De~Sande, T.~Gevers, and A.~W. Smeulders, ``Selective
  search for object recognition,'' \emph{International journal of computer
  vision}, vol. 104, no.~2, pp. 154--171, 2013.

\bibitem{sppnet}
K.~He, X.~Zhang, S.~Ren, and J.~Sun, ``Spatial pyramid pooling in deep
  convolutional networks for visual recognition,'' in \emph{European Conference
  on Computer Vision}.\hskip 1em plus 0.5em minus 0.4em\relax Springer, 2014,
  pp. 346--361.

\bibitem{visualize}
M.~D. Zeiler and R.~Fergus, ``Visualizing and understanding convolutional
  networks,'' in \emph{European conference on computer vision}.\hskip 1em plus
  0.5em minus 0.4em\relax Springer, 2014, pp. 818--833.

\bibitem{fcn}
J.~Long, E.~Shelhamer, and T.~Darrell, ``Fully convolutional networks for
  semantic segmentation,'' in \emph{IEEE Conference on Computer Vision and
  Pattern Recognition}, 2015, pp. 3431--3440.

\bibitem{instance2016}
J.~Dai, K.~He, and J.~Sun, ``Instance-aware semantic segmentation via
  multi-task network cascades,'' in \emph{IEEE Conference on Computer Vision
  and Pattern Recognition}, 2016, pp. 3150--3158.

\bibitem{instance2017}
Y.~Li, H.~Qi, J.~Dai, X.~Ji, and Y.~Wei, ``Fully convolutional instance-aware
  semantic segmentation,'' 2017.

\bibitem{mask_rcnn}
K.~He, G.~Gkioxari, P.~Dollar, and R.~Girshick, ``Mask r-cnn,'' \emph{IEEE
  International Conference on Computer Vision}, 2017.

\bibitem{cloud_cnn}
C.~Shi, C.~Wang, Y.~Wang, and B.~Xiao, ``Deep convolutional activations-based
  features for ground-based cloud classification,'' \emph{IEEE Geoscience and
  Remote Sensing Letters}, vol.~14, no.~6, pp. 816--820, 2017.

\bibitem{NBNN}
O.~Boiman, E.~Shechtman, and M.~Irani, ``In defense of nearest-neighbor based
  image classification,'' in \emph{IEEE Conference on Computer Vision and
  Pattern Recognition}, 2008, pp. 1--8.

\bibitem{C_R_are_one}
L.~Xie, R.~Hong, B.~Zhang, and Q.~Tian, ``Image classification and retrieval
  are one,'' in \emph{ACM on International Conference on Multimedia
  Retrieval}.\hskip 1em plus 0.5em minus 0.4em\relax ACM, 2015, pp. 3--10.

\bibitem{PWA}
J.~Xu, C.~Shi, C.~Qi, C.~Wang, and B.~Xiao, ``Part-based weighting aggregation
  of deep convolutional features for image retrieval,'' in \emph{AAAI}, 2018.

\bibitem{multiple_bow_ijcv}
H.~Jegou, M.~Douze, and C.~Schmid, ``Improving bag-of-features for large scale
  image search,'' \emph{International journal of computer vision}, vol.~87,
  no.~3, pp. 316--336, 2010.

\bibitem{multiple_bow_pami}
H.~Jegou, C.~Schmid, H.~Harzallah, and J.~Verbeek, ``Accurate image search
  using the contextual dissimilarity measure,'' \emph{IEEE Transactions on
  Pattern Analysis and Machine Intelligence}, vol.~32, no.~1, pp. 2--11, 2010.

\bibitem{soft_bow_cvpr}
J.~Philbin, O.~Chum, M.~Isard, J.~Sivic, and A.~Zisserman, ``Lost in
  quantization: Improving particular object retrieval in large scale image
  databases,'' in \emph{Computer Vision and Pattern Recognition, 2008. CVPR
  2008. IEEE Conference on}, 2008, pp. 1--8.

\bibitem{soft_bow_pami}
J.~C. van Gemert, C.~J. Veenman, A.~W. Smeulders, and J.~M. Geusebroek,
  ``Visual word ambiguity.'' \emph{IEEE Transactions on Pattern Analysis and
  Machine Intelligence}, vol.~32, no.~7, pp. 1271--83, 2010.

\bibitem{lclc}
J.~Wang, J.~Yang, K.~Yu, F.~Lv, T.~Huang, and Y.~Gong, ``Locality-constrained
  linear coding for image classification,'' in \emph{Computer Vision and
  Pattern Recognition (CVPR), 2010 IEEE Conference on}.\hskip 1em plus 0.5em
  minus 0.4em\relax IEEE, 2010, pp. 3360--3367.

\bibitem{AlexNet}
A.~Krizhevsky, I.~Sutskever, and G.~E. Hinton, ``Imagenet classification with
  deep convolutional neural networks,'' \emph{Communications of the Acm},
  vol.~60, no.~2, p. 2012, 2012.

\bibitem{GoogLeNet}
C.~Szegedy, W.~Liu, Y.~Jia, P.~Sermanet, S.~Reed, D.~Anguelov, D.~Erhan,
  V.~Vanhoucke, and A.~Rabinovich, ``Going deeper with convolutions,'' in
  \emph{Computer Vision and Pattern Recognition}, 2015, pp. 1--9.

\bibitem{ResNet}
K.~He, X.~Zhang, S.~Ren, and J.~Sun, ``Deep residual learning for image
  recognition,'' pp. 770--778, 2016.

\bibitem{AttentionNet}
F.~Wang, M.~Jiang, C.~Qian, S.~Yang, C.~Li, H.~Zhang, X.~Wang, and X.~Tang,
  ``Residual attention network for image classification,'' 2017.

\bibitem{SWIMCAT}
S.~Dev, Y.~H. Lee, and S.~Winkler, ``Categorization of cloud image patches
  using an improved texton-based approach,'' in \emph{IEEE International
  Conference on Image Processing}, 2015.

\bibitem{paris}
J.~Philbin, O.~Chum, M.~Isard, J.~Sivic, and A.~Zisserman, ``Lost in
  quantization: Improving particular object retrieval in large scale image
  databases,'' in \emph{IEEE Conference on Computer Vision and Pattern
  Recognition}.\hskip 1em plus 0.5em minus 0.4em\relax IEEE, 2008, pp. 1--8.

\bibitem{WAHRSIS}
S.~Dev, F.~M. Savoy, Y.~H. Lee, and S.~Winkler, ``Wahrsis: A low-cost
  high-resolution whole sky imager with near-infrared capabilities,'' vol.
  9071, no.~18, p. 90711L, 2014.

\bibitem{caffe}
Y.~Jia, E.~Shelhamer, J.~Donahue, S.~Karayev, J.~Long, R.~Girshick,
  S.~Guadarrama, and T.~Darrell, ``Caffe: Convolutional architecture for fast
  feature embedding,'' in \emph{ACM international conference on
  Multimedia}.\hskip 1em plus 0.5em minus 0.4em\relax ACM, 2014, pp. 675--678.

\bibitem{qe}
O.~Chum, J.~Philbin, J.~Sivic, M.~Isard, and A.~Zisserman, ``Total recall:
  Automatic query expansion with a generative feature model for object
  retrieval,'' in \emph{IEEE International Conference on Computer
  Vision}.\hskip 1em plus 0.5em minus 0.4em\relax IEEE, 2007, pp. 1--8.

\bibitem{IME}
J.~Xu, C.~Wang, C.~Qi, C.~Shi, and B.~Xiao, ``Iterative manifold embedding
  layer learned by incomplete data for large-scale image retrieval,''
  \emph{arXiv preprint arXiv:1707.09862}, 2017.

\bibitem{SanFrancisco}
D.~M. Chen, G.~Baatz, K.~K?ser, and S.~S. Tsai, ``City-scale landmark
  identification on mobile devices,'' in \emph{IEEE Conference on Computer
  Vision and Pattern Recognition}, 2011, pp. 737--744.

\bibitem{all_about_VLAD}
R.~Arandjelovic and A.~Zisserman, ``All about vlad,'' in \emph{IEEE Conference
  on Computer Vision and Pattern Recognition}, 2013, pp. 1578--1585.

\bibitem{three_thing}
------, ``Three things everyone should know to improve object retrieval,'' in
  \emph{IEEE Conference on Computer Vision and Pattern Recognition}.\hskip 1em
  plus 0.5em minus 0.4em\relax IEEE, 2012, pp. 2911--2918.

\bibitem{whitening}
H.~J¨¦gou and O.~Chum, ``Negative evidences and co-occurences in image
  retrieval: the benefit of pca and whitening,'' in \emph{European Conference
  on Computer Vision}, 2012, pp. 774--787.

\bibitem{ImageNet}
J.~Deng, W.~Dong, R.~Socher, and L.~J. Li, ``Imagenet: A large-scale
  hierarchical image database,'' in \emph{IEEE conference on computer vision
  and pattern recognition}, 2009, pp. 248--255.

\bibitem{Heinle}
A.~Heinle, A.~Macke, and A.~Srivastav, ``Automatic cloud classification of
  whole sky images,'' \emph{Atmospheric Measurement Techniques Discussions},
  vol.~3, no.~3, pp. 557--567, 2010.

\bibitem{LBP}
T.~Ojala, M.~Pietikainen, and T.~Maenpaa, ``Multiresolution gray-scale and
  rotation invariant texture classification with local binary patterns,'' in
  \emph{European Conference on Computer Vision}, 2000, pp. 404--420.

\bibitem{DCAF}
C.~Shi, C.~Wang, Y.~Wang, and B.~Xiao, ``Deep convolutional activations-based
  features for ground-based cloud classification,'' \emph{IEEE Geoscience and
  Remote Sensing Letters}, vol.~PP, no.~99, pp. 1--5, 2017.

\bibitem{LIBLINEAR}
R.~E. Fan, K.~W. Chang, C.~J. Hsieh, X.~R. Wang, and C.~J. Lin, ``Liblinear: A
  library for large linear classification,'' \emph{Journal of Machine Learning
  Research}, vol.~9, no.~9, pp. 1871--1874, 2012.

\end{thebibliography}

\end{document}